\documentclass[10pt,twocolumn,letterpaper]{article}

\usepackage[pagenumbers]{cvpr} 

\usepackage{graphicx}
\usepackage{amsmath}
\usepackage{amssymb}
\usepackage{booktabs}
\usepackage{multirow}
\usepackage{bbding}
\usepackage[pagebackref,breaklinks,colorlinks]{hyperref}

\usepackage[capitalize]{cleveref}
\crefname{section}{Sec.}{Secs.}
\Crefname{section}{Section}{Sections}
\Crefname{table}{Table}{Tables}
\crefname{table}{Tab.}{Tabs.}

\def\cvprPaperID{2317} 
\def\confName{CVPR}
\def\confYear{2023}

\begin{document}

\title{Clover \includegraphics[width=12pt]{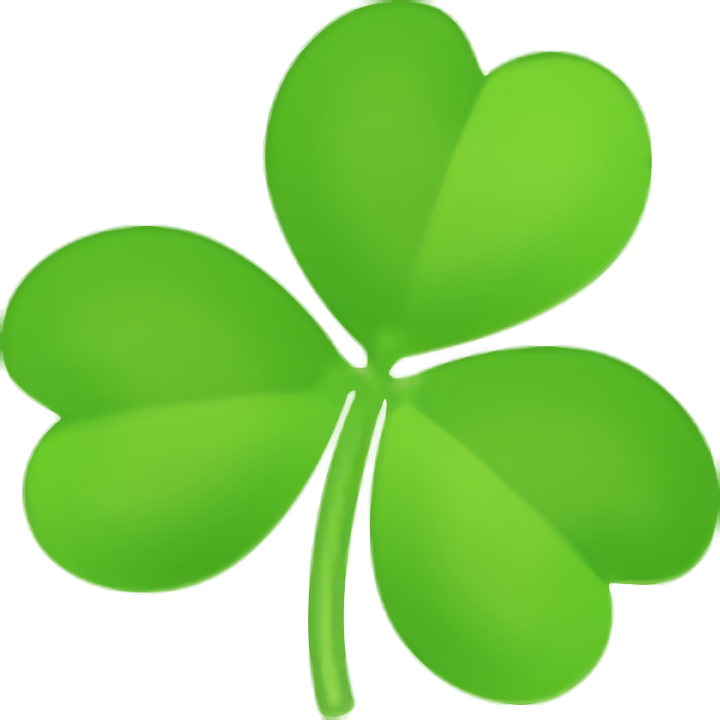}: Towards A Unified Video-Language Alignment and Fusion Model   }

\author{
  Jingjia Huang$^{\diamondsuit}$\thanks{Equal Contribution.}
  \and
  Yinan Li$^{\heartsuit*}$\thanks{Work done when interning at ByteDance Inc.}
  \and
  Jiashi Feng$^\diamondsuit$
  \and
  Xinglong Wu$^\diamondsuit$
  \and
  Xiaoshuai Sun$^\heartsuit$\thanks{Corresponding Author.}
  \and
  Rongrong Ji$^\heartsuit$
  \\[8pt]
  $^\heartsuit$Media Analytics and Computing Lab, Department of Artificial Intelligence,
  \\
  School of Informatics, Xiamen University,  361005, China
  \\
  $^\diamondsuit$ByteDance Inc, 100043, China
  \\
  yinanlee@stu.xmu.edu.cn, \{xssun, rrji\}@xmu.edu.cn
  \\
  \{huangjingjia, 
jshfeng,
wuxinglong\}@bytedance.com
}

\maketitle

\begin{abstract}
Building a universal Video-Language model for solving various video understanding tasks (\emph{e.g.}, text-video retrieval, video question answering) is an open challenge to the machine learning field. Towards this goal, most recent   works build the model by stacking  uni-modal and cross-modal feature encoders and train it with  pair-wise contrastive  pre-text tasks. Though offering attractive generality,  the resulted models  have to compromise between efficiency and performance. They mostly adopt different architectures to deal with different downstream tasks.
We find this is because the pair-wise training cannot   well  \emph{align} and \emph{fuse} features from different modalities. We then introduce \textbf{Clover}\textemdash  a Correlated Video-Language pre-training method\textemdash towards a universal Video-Language model for solving multiple video understanding tasks with neither performance nor efficiency compromise. It improves  cross-modal feature alignment and fusion via a novel tri-modal alignment pre-training task. Additionally, we propose to enhance the tri-modal alignment via incorporating learning from semantic masked samples and a new pair-wise ranking loss.
Clover establishes new state-of-the-arts on multiple downstream tasks, including three retrieval tasks  for both zero-shot and fine-tuning settings,  and eight video question answering tasks. 
Codes and pre-trained models will be released at \url{https://github.com/LeeYN-43/Clover}.
\end{abstract}

\section{Introduction}
Video-Language pre-training (VidL)   aims to learn generalizable multi-modal models  from large-scale video-text samples so as to better solve various challenging Video-Language understanding tasks, such as text-video retrieval \cite{maharaj2017dataset, chen2011collecting, anne2017localizing, xu2016msr} and video question answering \cite{jang2017tgif, torabi2016learning, xu2017video}. Recent studies~\cite{lei2021less, li2022alignandprompt, fu2021violet, ge2022bridging, yang2021just, wang2022all, Xue_2022_CVPR, zellers2021merlot} have shown that VidL leads to significant performance improvement  and achieves state-of-the-art results on various downstream text-video retrieval and video question answering (VQA) benchmarks.

Though achieving encouraging performance, existing VidL models mostly adopt different architectures to deal with different downstream tasks.
For the text-video retrieval tasks, they \cite{bain2021frozen, gabeur2020multi, ging2020coot, liu2019use, miech2020end, ge2022bridging} typically use  two individual uni-modal encoders for processing video and text data separately, for the sake of retrieval efficiency. 
While for video question answering tasks, the models usually adopt the multi-modal joint encoder design to learn the association and interaction of different modalities.
\begin{figure*}[t]
  \centering
  \includegraphics[width=1.9\columnwidth]{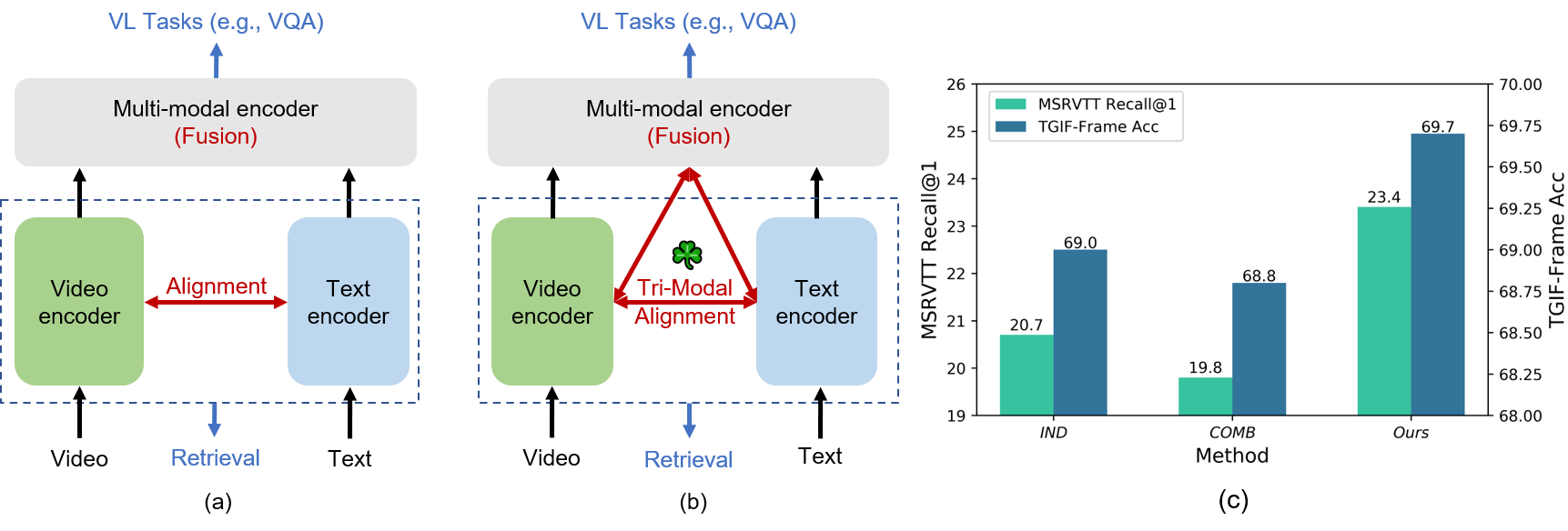}
  \caption{{Different attempts on achieving a universal VidL model}. (a) Naive combination by stacking the cross-modal encoder upon the uni-modal encoders with weak association (\emph{COMB}). (b) Our proposed Clover \includegraphics[width=8pt]{figure/shamrock-emoji.png} pre-trained model that better correlates the cross-modal encoder and the uni-modal encoders via Tri-Modal Alignment. (c) Comparisons of  the transfer  performance on MSRVTT \cite{yu2018joint} (zero-shot) and TGIF-FrameQA~\cite{jang2017tgif} datasets.  Benefiting from the proposed tri-modal alignment, our proposed universal model outperforms both the individually designed models (\emph{IND}) and \emph{COMB}.}
   \vspace{-1\baselineskip}
   \label{fig:fig1}
\end{figure*}

Building a unified model capable of solving various Video-Language tasks is a long-standing challenge for machine learning research. A few recent works \cite{fu2021violet, li2022alignandprompt} attempt to learn a unified VidL model for both tasks, which uses the multi-modal encoder to conduct text-video retrieval. 
However, the model requires an exhaustive pair-wise comparison between the query texts and gallery videos. Given $N$ text queries and $M$ category videos, the computation complexity of the multi-modal encoder model would be $O(NM)$, which makes it infeasible for large-scale video-text retrieval applications.
Another straightforward  solution is to simply combine the uni-modal  and   multi-modal encoders (Fig.~\ref{fig:fig1} (a)), and perform the retrieval and VQA tasks through the uni-modal   and multi-modal encoders respectively. 
Its computation complexity   for retrieval tasks is only $O(N+M)$. However, the experiment results in Fig.\ref{fig:fig1} (c) show that, without a carefully designed correlating mechanism between these two types of encoders, the simple combination \emph{i.e., COMB} yields a compromised performance compared with the models \emph{i.e. IND} individually designed for retrieval and VQA.

In this work, we aim to address  the above issues and  build a unified pre-training model that attains high efficiency and performance simultaneously. We observe: (i) well aligning the features of the video and text from the same data pair is important for text-video matching; (ii) effectively fusing video and text features into unified representations is critical for video-text understanding. However, existing pre-training strategies that rely on either simple supervised or contrastive pre-text tasks hardly achieve promising feature alignment and fusion capability simultaneously. Motivated by these observations, we develop a new VidL method from these two aspects.

Specifically, we propose Correlated Video-Language pre-training (Clover), a VidL method that not only unifies Video-Language alignment and fusion, but also makes them mutually boosted.
The fused multi-modal representation contains richer context information than the uni-modal representations \cite{ge2022bridging,Luo2020UniVL}. 
As an intermediate modality between video and text, the multi-modal representations are good anchors for cross-modality alignment. 
Meanwhile, keeping the fused representation closer to the uni-modal representation containing consistent semantic information and away from the inconsistent one will enhance the learning of semantic information in the fused modality.
Therefore, we propose the Tri-Modal Alignment (TMA) to get Video-Language alignment and fusion mutually boosted, which takes the alignment between the multi-modal representation and text/video representations as an auxiliary objective. We note that since the tri-modal alignment is well compatible with the classical pre-training tasks  \cite{devlin2019bert,bao2021beit, Luo2020UniVL} \emph{e.g.,} Masked Language Modeling, its  computation overhead is negligible.
To help the model maintain fine-grained discriminative capability while improving its generalizability, we further introduce a pair-wise ranking loss that urges the model to be aware of the concept missing in masked samples compared to original samples.

Extensive experiments are conducted on multiple downstream tasks, including three retrieval tasks with different experimental setups (\emph{i.e.} zero-shot and fine-tune) and eight video question answering tasks. The  results demonstrate that Clover is able to get the cross-modal fusion and alignment capability mutually improved, and consistently outperforms current SOTAs on various downstream tasks.  It achieves an average performance improvement of 4.9\% and 8.7\% Recall@10 score on the zero-shot and fine-tune settings of the three downstream retrieval datasets, while the average accuracy improvement over current SOTAs is 2.3\% on the eight video question answering datasets.

In summary, we make the following contributions: (1) we introduce Clover, a pre-training method achieving unified Video-Language alignment and fusion model that can be easily transferred to various downstream video understanding tasks while attaining both high efficiency and performance; 
(2)  we propose a novel tri-modal alignment pre-training task, which correlates the uni-modal encoder and multi-modal encoder to get them mutually boosted.

\section{Related Work}
\noindent \textbf{Video-Language pre-training for text-video retrieval}. Existing pre-training methods for text-video retrieval can be mainly divided into two categories. The first kind of method employs two individual encoders to embed video and text, and project them into a common latent space \cite{bain2021frozen, wang2022object, ge2022bridging, liu2019use, xu2021videoclip, miech2020end, gabeur2020multi, ging2020coot, rouditchenko2021avlnet, yang2021taco}. Then, they leverage a contrastive objective \cite{van2018representation}  to align the paired representations. Thanks to their high efficiency, this category of methods is widely used in real-world applications. The other kind of method utilizes a joint multi-modal encoder to learn comprehensive representation for a given video-text pair, and predict whether the video and text are matched or not via a binary classifier \cite{fu2021violet, lei2021less, li2020hero, sun2019videobert, xu2021vlm, zhu2020actbert}.
Despite the good performance they achieved, they need to exhaustively pair all the videos and texts, and feed them into the model during inference, making them highly inefficient.

\noindent  \textbf{Video-Language pre-training for video question answering.} 
Video question answering aims to automatically answer natural language questions given a context video, which requires a joint understanding of both the video and language. Previous works mainly focus on designing better spatio-temporal attention networks \cite{xu2017video, jang2017tgif, li2019beyond, zha2019spatiotemporal, liu2021hair, fan2019heterogeneous} and question-video relation networks \cite{kim2019progressive, le2020hierarchical, huang2020location, park2021bridge} to extract more accurate multi-modal representations. Recently, many works leverage transformer-based models to build an additional multi-modal encoder upon the uni-modal encoders \cite{seo2021look, fu2021violet, li2022alignandprompt} or input image patches and word tokens  \cite{zhu2020actbert, yang2021just} directly to fuse cross-modality features. The multi-modal encoder is typically driven by the pre-training tasks \emph{e.g.} Masked Language Modeling \cite{devlin2019bert} and Visual-Text Matching \cite{tan2019lxmert}, to perform token-level cross-modal fusion.

\noindent \textbf{Video-Language pre-training for multiple downstream tasks.}
There are a few recent works trying to construct a unified pre-trained model for multiple downstream tasks. ActBert \cite{zhu2020actbert}, clipBert \cite{lei2021less}, VIOLET \cite{fu2021violet} and ALPRO \cite{li2022alignandprompt} utilize a cross-modal encoder for both retrieval and VQA tasks. They take video-text retrieval as a binary classification task and construct the classifier upon a visual-text matching head, which brings about extreme computation overload. While All-in-One \cite{wang2022all} and HD-VILA \cite{Xue_2022_CVPR} proposes a transformer-based unified backbone architecture for joint-modal representation learning, they require a large amount of data for pre-training. HERO \cite{li2020hero} and VLM \cite{xu2021vlm} design a task-agnostic model that can be finetuned on multiple downstream tasks. However, they are not end-to-end trainable and need pre-extract video features, which limits the performance of the model.

Unlike the aforementioned methods, our 
Clover is a fully end-to-end VidL model that can be easily transferred to various downstream tasks while attaining both high efficiency and performance.

\section{Method}
\label{sec:method}

\subsection{Motivation and Overview}
Recent years have witnessed a significant progress in Video-Language pre-training. Nevertheless, building a universal Video-Language model for solving various video understanding tasks (\emph{e.g.}, text-video retrieval, video question answering) is still a challenging problem. The key to the problem lies in how to unify the representation learning for cross-modal alignment and fusion efficiently. 

Cross-modal \emph{alignment} aims to learn projection functions $f(\cdot)$ and $g(\cdot)$ that project videos and texts into a common embedding space where the similarity between the video and text with consistent semantic information is maximized and otherwise minimized. Taking a video ${V}$ as the anchor, the learning objective of cross-modal alignment is formulated as:
\begin{equation}
    \arg\max_{f,g}\left[\mathrm{s}\left(f(V), g(T^+)\right) - \mathrm{s}\left(f(V), g(T^-)\right)\right],
    \label{eq:cross-modal alignment}
\end{equation}
where $T^+, T^-$ represent texts that are semantically consistent and inconsistent with $V$, respectively. The function $\mathrm{s}(\cdot, \cdot)$ measures the dot-product similarity between two embeddings. A model with strong cross-modal alignment capability is desired in the video-retrieval task, which requires matching the semantically aligned videos and texts.

Cross-modal \emph{fusion} aims to integrate the correlation and interaction carried by the video and text modalities into a unified multi-modal embedding. Specifically, it can be formulated as learning a function $Fusion(\cdot, \cdot)$ that takes inputs of different modals and outputs a unified representation $M = Fusion(V, T)$, which is then used to solve downstream tasks like VQA.

Existing pre-training strategies rely on either simple supervised or contrastive pre-text tasks that   hardly achieve feature alignment and fusion simultaneously. To better correlate the cross-modal alignment and fusion, we propose the Correlated Video-Language pre-training method (Clover) with three key pre-training innovations (Sec.~\ref{sec:pre-training task}): tri-modal alignment; pair-wise ranking loss; and semantic enhanced masked language modeling. The classical masked language modeling task is also incorporated into our method, which is able to boost the generalizability of the model as well as the interaction between visual and language.

\begin{figure*}[t]
\vspace{-3\baselineskip}
  \centering
  \includegraphics[width=0.9\linewidth]{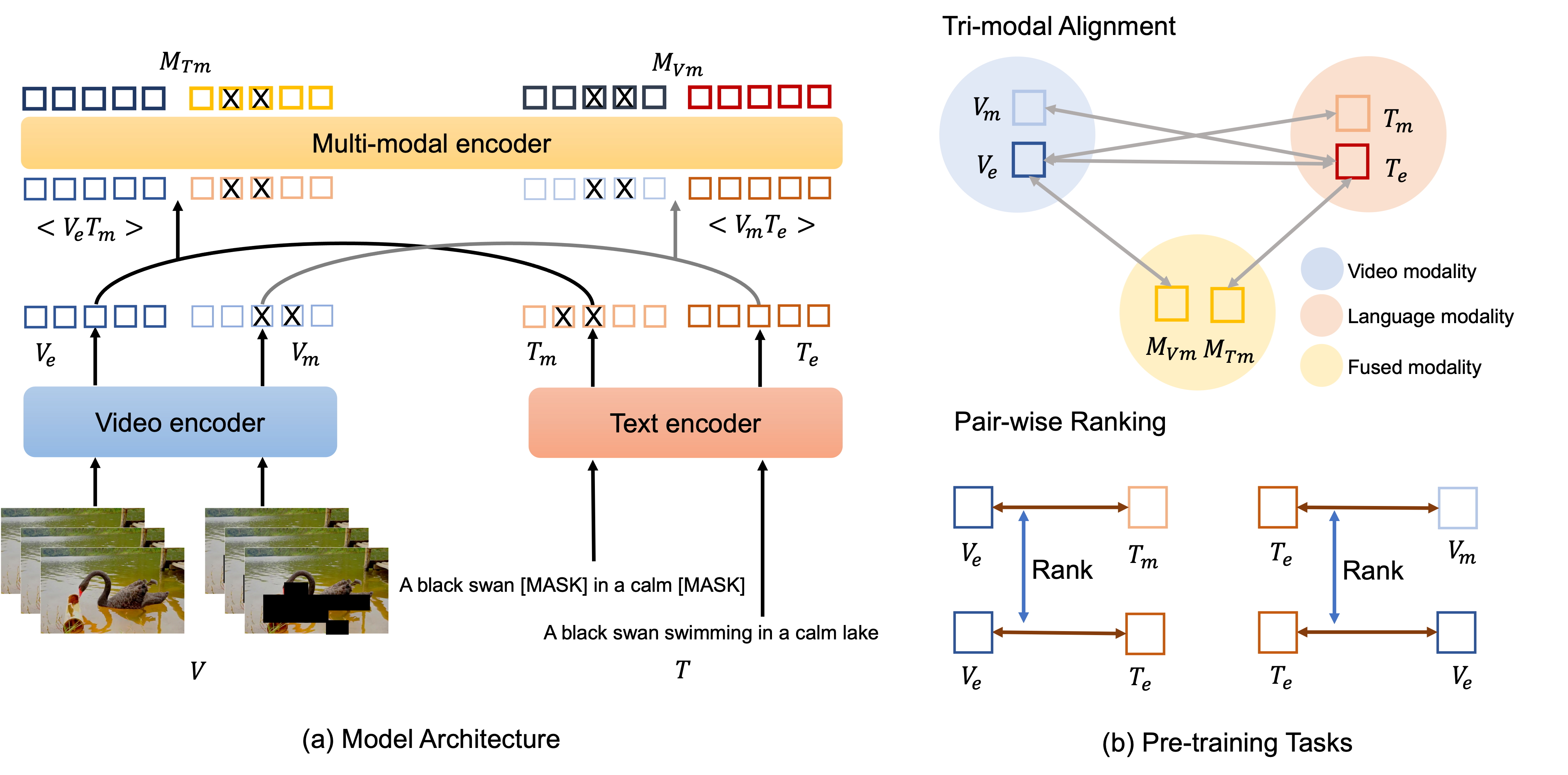}
  \caption{Overview of Clover \includegraphics[width=8pt]{figure/shamrock-emoji.png}. (a) Model architecture. (b) Tri-modal alignment pre-training task with masked samples and the pair-wise ranking. ``$\times$'' indicates the embedding of the token [MASK]. For more detail on model architecture and pre-training tasks refers to Sec. \ref{sec:method}. }
  \vspace{-1\baselineskip}
   \label{fig:method}
\end{figure*}

\noindent  \textbf{Architecture}. We briefly explain the  architecture of Clover here, which is shown in Fig.~\ref{fig:method} as well. It  consists of three components: a video encoder, a text encoder and a multi-modal encoder. Following \cite{fu2021violet}, we use VideoSwin Transformer~\cite{liu2022video} as our video encoder. Given a video $V$, it outputs a sequence of video embeddings: $V_e = \{v_1, ..., v_{K}\} \subset \mathbb{R}^D$, where $K$ is the number of flattened patches. The text encoder of Clover is a 12-layer bidirectional transformer encoder model \cite{devlin2019bert}. Given an input text sentence $T$, the encoder outputs an embedding sequence $T_e = \{t_{\text{cls}}, t_1, ..., t_{L-1}\} \subset \mathbb{R}^D$, where  $t_{\text{cls}}$ indicates the embedding of the [CLS] token. We employ a 3-layer bidirectional Transformer encoder to learn the fused cross-modal representation of a video-text pair. The multi-modal encoder takes the concatenation of video and text embeddings as input, and outputs the fused multi-modal embeddings $M_e = \{m_{v_1}, ..., m_{v_K}, m_{\mathrm{CLS}}, m_{t_1}, ..., m_{t_{L-1}}\}$, where $m \in \mathbb{R}^D$. 

\subsection{Tri-Modal Alignment} \label{sec:pre-training task}

Existing VidL pre-training methods typically align the embeddings of data from different modalities through pair-wise contrastive learning. Differently, we propose a Tri-Modal Alignment (TMA) learning task that not only enforces the video and text modalities to be aligned but also encourages them to well align with the third modality, \emph{i.e.,} their fusion modality.
In TMA, as an intermediate modality between video and text, the fused multi-modal representations act as anchors for cross-modality alignment, which reduces the difficulty of the direct alignment between video and text. Meanwhile, it keeps the fused representation closer to the uni-modal representation containing consistent semantic information and away from the others, so as to enhance the learning of semantic information in the fused representation.

As shown in Fig.~\ref{fig:method} (a), given an input video-text pair $\left\langle V, T \right\rangle$ and their embeddings  $\left\langle V_e, T_e \right\rangle$ from the uni-modal encoder, we mask some regions in $V$ to make a masked  video with embedding $V_m$, and similarly mask some tokens in $T$ to make a   masked  text embedding  $T_m$. Then, we pair the incomplete samples with the complete samples as $\langle V_m,T_e \rangle$ and $\langle V_e,T_m \rangle$. We adopt multi-modal encoder to get fused multi-modal embedding sequence $M_{V_m}$ and $M_{T_m}$ of $\langle V_m, T_e \rangle$ and $\langle V_e, T_m \rangle$, respectively. We denote the embedding of [CLS] token in $M_{V_m}$ and $M_{T_m}$ as $M_{{V_{m}f}}$ and $M_{{T_{m}f}}$.

 For the convenience of description, we use superscripts $i$ and $j$ to index the data sample, and introduce TMA from the perspective of video and language, respectively. Firstly, for a video representation $V_e^i$, besides its associated text embedding $T_e^i$ and $T_m^i$, we also consider its associated fusion embedding $M_{{V_{m}f}}^i$ as its positive pair\textemdash that should be pushed closer within the embedding space. To align these tri-modal representations, we propose a novel exclusive-NCE loss computed within a batch of $B$ samples as follows:
\begin{equation}
\begin{aligned}
     L_{v} &= - \sum_{i=1}^B \left [
    \log \frac{e^{s(V_e^i, T^i_{{e}}) / \tau}}
    {e^{s(V_{e}^i, T^i_{{e}} ) / \tau} + Z } 
    + \log \frac{e^{s(V_e^i, T^i_{ {m}}) / \tau}} 
    {e^{s(V_{e}^i, T^i_{{m}} ) / \tau} + Z } \right.\\ 
    & \left. + \log \frac{e^{s(V_e^i, M^i_{{V_{m}f}}) / \tau}} 
    {e^{s(V_{e}^i, M^i_{{V_{m}f}} ) / \tau} + Z } \right ]
    , where\\
 Z &= \sum_{j \neq i}^B \left[ e^{s(V_{e}^i, T_e^j)/ \tau} + e^{s(V_{e}^i, T_{{m}}^j)/ \tau} + e^{s(V_{e}^i, M_{{V_{m}f}}^j)/ \tau}\right] .
    \end{aligned}
    \label{eq:exclusive-NCE v2t}
\end{equation}
Here $\tau$ is a temperature scalar, and $\mathrm{s}(\cdot, \cdot)$ is the dot-product similarity function to measure the degree of alignment between different modalities. In each term of Eq.~\eqref{eq:exclusive-NCE v2t}, we exclude the other positive pairs when performing contrastive learning on one positive pair. 
Thus the intra-suppression among positive pairs is effectively avoided, and  the video-/text-/fusion-modal representations are better aligned. For example, in the first term of Eq.~\eqref{eq:exclusive-NCE v2t}, $e^{s(V_e^i, M_{V_{m}f}^{i}) / \tau}$ and $e^{s(V_e^i, T_m^{i}) / \tau}$ are excluded from the denominator, which prevents the model from yielding the sub-optimal solution that gets  $T_{e}^i$ closer to $V_{e}^i$ while pushes $T_m^{i}$ and $M_{V_{m}f}^{i}$ away from $V_{e}^i$.
Meanwhile, to align the video to the text and fusion modalities, we have the following objective: 
\begin{equation}
\begin{aligned}
    L_{v'} &= - \sum_{i=1}^B \left[
    \log \frac{e^{s(T^i_{{e}}, V_{e}^i) / \tau}}{\sum_{j=1}^B e^{s(T^i_{{e}}, V_e^j)/ \tau}} + \log \frac{e^{s(T^i_{{m}}, V_{e}^i) / \tau}}{\sum_{j=1}^B e^{s(T^i_{{m}}, V_e^j)/ \tau}}\right.\\
    &\left.+ \log \frac{e^{s(M^i_{{V_{m}f}}, V_{e}^i) / \tau}}{\sum_{j=1}^B e^{s(M^i_{{V_{m}f}}, V_e^j)/ \tau}} \right].
    \label{eq:exclusive-NCE t2v}
    \end{aligned}
\end{equation}
The above two losses  \eqref{eq:exclusive-NCE v2t} and \eqref{eq:exclusive-NCE t2v} are summed up to give the tri-modal alignment objective w.r.t.\ the visual modality:  $L_{V} = L_{v} + L_{v'}$. 
Similarly, from the perspective of text modality, we define the tri-modal alignment objective $L_T=L_{t}+L_{t'}$ w.r.t.\ the text modality. \footnote{For the detailed formulation of $L_T$, please refer to the supplementary material.} 
Then, we obtain the final loss of tri-modal alignment $L_{TmA} = L_V + L_T$. 

To get the masked samples for TMA, we conduct masking over the videos and texts to form masked samples using the following two strategies. 

\noindent \textbf{Video-block masking strategy}. 
 Objects in a video often appear at similar spatial locations across consecutive frames, which compose a tube area in the spatial-temporal domain. In order to construct incomplete video samples with less information leakage, we extend the block-wise mask \cite{bao2021beit} to video via performing a block-wise mask at the same position in all video frames. In this paper, we randomly replace 20\% patches with a learnable mask token to obtain $V_m$.

\noindent  \textbf{Semantic text masking strategy}. The detailed semantic information in the text basically lies in the verb, noun and adjective words. To facilitate learning their representations, we construct incomplete text samples by randomly masking some verbs, nouns and adjectives in the sentence. Specifically, given a sentence, we use a part-of-speech tagger \cite{loper2002nltk} to tag each word. Then we pick the verb phrases and nouns, and replace 30\% of them with a special [MASK] token. Note that to avoid drastic semantic changes, we do not mask the auxiliary verb like \emph{have, should, will, would}, etc. With the masking strategy, when some key elements are masked from the  texts, it would form an expression carrying partial information rather than completely changing the information expressed.

\subsection{Training Objective}
\noindent  \textbf{Pair-wise ranking}.
In TMA, we take the masked pairs $\left\langle V_e, T_m \right\rangle$ and $\left\langle V_m, T_e \right\rangle$ as positive pairs, considering that masked samples would still carry partial meaningful information though some others are missed. For instance, as shown in Fig.~\ref{fig:method} (a), given the text ``A black swan swimming in a calm lake'', ``swimming'' and ``lake'' are masked with [MASK] tokens, generating the masked pair $\langle V_{e},T_{m} \rangle$. Although some   information is removed, the masked sentence still contains other concepts that make it partially matched with the video, such as the ``black swan''. Therefore, in contrastive learning, $V_m$ and $T_m$ can be considered pseudo positive candidates for $T$ and $V$. However, compared to the original pair $\left\langle V_e, T_e \right\rangle$, we consider that the semantic consistency in the masked pairs should be weaker. Based on this prior, we further propose a pair-wise ranking loss:
\begin{equation}
\begin{aligned}
    L_{rank} &=  \max\left( 0, -\left(\text{sim}(V_e, T_e) / \tau - \text{sim}(V_e, T_m) / \tau \right) + \lambda \right)\\
    &+ \max\left(0, -\left(\text{sim}(V_e, T_e) / \tau - \text{sim}(V_m, T_e) / \tau \right) + \lambda \right), \\
\end{aligned}
\label{eq:ranking loss}
\end{equation}
where $\lambda > 0$ is a margin hyper-parameter. Eq.~\eqref{eq:ranking loss} urges the model to be aware of the gap of semantic consistency between $\left\langle V_e, T_m \right\rangle / \left\langle V_m, T_e \right\rangle$ and $\left\langle V_e, T_e \right\rangle$ brought about by concepts missing in the masked pairs. With the pair-wise ranking objective, our model is able to maintain fine-grained perceptual capability while improving its generalizability. 

\noindent  \textbf{Semantic enhanced masked language modeling}. Masked language modeling (MLM) is a classical pre-training task in VidL, which promotes the interaction between different modalities in the cross-modal encoder. Combining the classical MLM with our semantic text masking strategy, we present the Semantic Enhanced Masked Language Modeling task, which facilitates the representation learning on the key concepts, \emph{i.e.,} verb, noun and adjective words. Besides, considering the class-imbalance between different words, we use focal loss \cite{lin2017focal} instead of the traditional cross-entropy loss to improve the MLM loss. We apply the MLM loss to the reconstruction of the masked text tokens \emph{i.e.,} $m_{t_j} \in M_{T_m}$, where $t_j=[Mask]$. The MLM loss is defined as:
\begin{equation}
\begin{aligned}
    L_{mlm} &= -\frac{1}{B} \sum_{i=1}^{B} \sum_{m_{t_j} \in  M^{i}_{T_{m}}} [(1 - p^{i}_{t_j})^\gamma p^{i}_{t_j}]
    \label{eq:mlm_loss}
\end{aligned}
\end{equation}
where $p^{i}_{t_j}$ denote the predicted probability distribution of the masked token $t_j$ in i$th$ sentence in the batch, and $\gamma$ is a hyper-parameter. Finally, the overall pre-training objective of Clover is:
\begin{equation}
    L = L_{TmA} + L_{rank} + L_{mlm}.
    \label{eq:all_loss}
\end{equation}

\begin{table*}[t]
\centering
\resizebox{0.98\textwidth}{!}{
\begin{tabular}{cccccccccccccc}
\toprule
\multicolumn{1}{l|}{\multirow{2}{*}{Method}} & \multicolumn{1}{c|}{\multirow{2}{*}{Pre-training dataset}} & \multicolumn{4}{c|}{DiDeMo} & \multicolumn{4}{c|}{MSRVTT} & \multicolumn{4}{c}{LSMDC} \\ \cline{3-14} 
\multicolumn{1}{l|}{} & \multicolumn{1}{c|}{} & R@1 & R@5 & R@10 & \multicolumn{1}{c|}{MedR} & R@1 & R@5 & R@10 & \multicolumn{1}{c|}{MedR} & R@1 & R@5 & R@10 & MedR \\ \midrule
\multicolumn{14}{c}{Fine-tune} \\ \midrule
\multicolumn{1}{l|}{clipBert \cite{lei2021less}} & \multicolumn{1}{c|}{COCO\cite{chen2015microsoft}, VG\cite{krishna2017visual}} & 20.4 & 48.0 & 60.8  & \multicolumn{1}{c|}{6} & 22.0 & 46.8 & 59.9 & \multicolumn{1}{c|}{6} & - & - & - & - \\
\multicolumn{1}{l|}{Frozen \cite{bain2021frozen}} & \multicolumn{1}{c|}{W2M+C3M} & 31.0 & 59.8 & 72.4 & \multicolumn{1}{c|}{3} & 31.0 & 59.5 & 70.5 & \multicolumn{1}{c|}{3} & 15.0 & 30.8 & 39.8 & 20 \\
\multicolumn{1}{l|}{VIOLET \cite{fu2021violet}} & \multicolumn{1}{c|}{W2M+C3M+Y180M}& 32.6 & 62.8 & 74.7 & \multicolumn{1}{c|}{-} & 34.5 & 63.0 & 73.4 & \multicolumn{1}{c|}{-} & 16.1 & 36.6 & 41.2 & - \\
\multicolumn{1}{l|}{HD-VILA \cite{Xue_2022_CVPR}} & \multicolumn{1}{c|}{HDV100M} & 28.8 & 57.4 & 69.1 & \multicolumn{1}{c|}{4} & 35.6 & 65.3 & 78.0 & \multicolumn{1}{c|}{3} & 17.4 & 34.1 & 44.1 & 15 \\

\multicolumn{1}{l|}{ALPRO \cite{li2022alignandprompt}} & \multicolumn{1}{c|}{W2M+C3M} & 35.9 & 67.5 & 78.8 & \multicolumn{1}{c|}{3} & 33.9 & 60.7 & 73.2 & \multicolumn{1}{c|}{3} & - & - & - & - \\
\multicolumn{1}{l|}{TMVM \cite{lin2022text}} & \multicolumn{1}{c|}{W2M+C3M} & 36.5 & 64.9 & 75.4 & \multicolumn{1}{c|}{3} & 36.2 & 64.2 & 75.7 & \multicolumn{1}{c|}{3} & 17.8 & 37.1 & 45.9 & 13.5 \\
\multicolumn{1}{l|}{All-in-1 \cite{wang2022all}} & \multicolumn{1}{c|}{W2M+H100M} & 32.7 & 61.4 & 73.5 & \multicolumn{1}{c|}{-} & 37.9 & 68.1 & 77.1 & \multicolumn{1}{c|}{-} & - & - & - & - \\
\multicolumn{1}{l|}{OA-Trans \cite{wang2022object}} & \multicolumn{1}{c|}{W2M+C3M}& 34.8 & 64.4 & 75.1& \multicolumn{1}{c|}{3}  & 35.8 & 63.4 & 76.5 & \multicolumn{1}{c|}{3} & 18.2 & 34.3 & 43.7 & 18.5 \\
\multicolumn{1}{l|}{MILES \cite{ge2022miles}} & \multicolumn{1}{c|}{W2M+C3M}& 36.6 & 63.9 & 74.0 & \multicolumn{1}{c|}{3}  & 37.7 & 63.6 & 73.8 & \multicolumn{1}{c|}{3} & 17.8 & 35.6 & 44.1 & 15.5 \\ 
\multicolumn{1}{l|}{MCQ \cite{ge2022bridging}} & \multicolumn{1}{c|}{W2M+C3M} & 37.0 & 62.2 & 73.9  & \multicolumn{1}{c|}{3} & 37.6 & 64.8 & 75.1 & \multicolumn{1}{c|}{3} & 17.9 & 35.4 & 44.5 & 15 \\ 

\midrule
\multicolumn{1}{l|}{Clover (ours) } & \multicolumn{1}{c|}{W2M+C3M} & \textbf{50.1} & \textbf{76.7} & \textbf{85.6} & \multicolumn{1}{c|}{\textbf{1}} & \textbf{40.5} & \textbf{69.8} & \textbf{79.4} & \multicolumn{1}{c|}{\textbf{2}} & \textbf{24.8} & \textbf{44.0} & \textbf{54.5} & \textbf{8} \\ \midrule

\multicolumn{14}{c}{Zero-shot} \\ \midrule
\multicolumn{1}{l|}{MIL-NCE \cite{miech2020end}} & \multicolumn{1}{c|}{H100M} & - & - & - & \multicolumn{1}{c|}{-}  & 9.9 & 24.0 & 32.4 & \multicolumn{1}{c|}{29.6}& - & - & - & - \\
\multicolumn{1}{l|}{Frozen \cite{bain2021frozen}} & \multicolumn{1}{c|}{W2M+C3M}  & 21.1 & 46.0 & 56.2 & \multicolumn{1}{c|}{7} & 18.7 & 39.6 & 51.6 & \multicolumn{1}{c|}{10} & 9.3 & 22.0 & 30.1 & 51.0 \\
\multicolumn{1}{l|}{VIOLET \cite{fu2021violet}} & \multicolumn{1}{c|}{W2M+C3M+Y180M} & 23.5 & 49.8 & 59.8 & \multicolumn{1}{c|}{-} & 25.9 & \textbf{49.5} & 59.7 & \multicolumn{1}{c|}{-} & - & - & - & - \\
\multicolumn{1}{l|}{ALPRO \cite{li2022alignandprompt}} & \multicolumn{1}{c|}{W2M+C3M}  & 23.8 & 47.3 & 57.9 & \multicolumn{1}{c|}{6} & 24.1 & 44.7 & 55.4 & \multicolumn{1}{c|}{8} & - & - & - & - \\
\multicolumn{1}{l|}{OA-Trans \cite{wang2022object}} & \multicolumn{1}{c|}{W2M+C3M}  & 23.5 & 50.4 & 59.8 & \multicolumn{1}{c|}{6} & 23.4 & 47.5 & 55.6 & \multicolumn{1}{c|}{8} & - & - & - & - \\
\multicolumn{1}{l|}{MILES \cite{ge2022miles}} & \multicolumn{1}{c|}{W2M+C3M}  & 27.2 & 50.3 & 63.6 &  \multicolumn{1}{c|}{5} & 26.1 & 47.2 & 56.9 & \multicolumn{1}{c|}{7} & 11.1 & 24.7 & 30.6 & 50.7 \\ 
\multicolumn{1}{l|}{MCQ \cite{ge2022bridging}} & \multicolumn{1}{c|}{W2M+C3M}  & 25.6 & 50.6 & 61.1 &  \multicolumn{1}{c|}{5} & 26.0 & 46.4 & 56.4 & \multicolumn{1}{c|}{7} & 12.2 & 25.9 & 32.2 & 42 \\ 
\midrule
\multicolumn{1}{l|}{Clover (ours)} & \multicolumn{1}{c|}{W2M+C3M}  & \textbf{29.5} & \textbf{55.2} & \textbf{66.3} & \multicolumn{1}{c|}{\textbf{4}} & \textbf{26.4} & \textbf{49.5} & \textbf{60.0} & \multicolumn{1}{c|}{\textbf{6}} & \textbf{14.7} & \textbf{29.2} & \textbf{38.2} & \textbf{24} \\ \bottomrule
\end{tabular}}
\caption{Text-to-video retrieval performance  comparison under {fine-tune} and {zero-shot} setups. Here {higher} R@k (Recall K) and {lower} MedR (Median Recall) indicate better performance. W2M, C3M, H100M, HDV100M, Y180M are short for WebVid2M\cite{bain2021frozen}, CC3M\cite{sharma2018conceptual}, HowTo100M\cite{miech2019howto100m},
HD-VILA-100M\cite{Xue_2022_CVPR},
YT-Temporal-180M\cite{zellers2021merlot}, respectively.}
\vspace{-0.2cm}
\label{tab:retrieval_SOTA}
\end{table*}

\section{Experiments}
\subsection{Experiment Setup}
\noindent  \textbf{Pre-training datasets}.
Following recent work \cite{bain2021frozen, li2022alignandprompt, ge2022bridging}, we jointly pre-train our Clover on a video dataset WebVid2M \cite{bain2021frozen} with 2.5M video-text pairs and an image dataset Google Conceptual Captions (CC3M) \cite{sharma2018conceptual} with 3.3M image-text pairs (we only obtain 2.8M image-text pairs in CC3M due to image url broken). During pre-training, we treat image data as one frame video data.

\noindent  \textbf{Downstream tasks}. We evaluate our proposed models on the following downstream tasks. (a) \textbf{Text-to-Video Retrieval} on MSRVTT \cite{xu2016msr}, LSMDC  \cite{maharaj2017dataset} and DiDeMo \cite{anne2017localizing}. For DiDeMo, we follow \cite{lei2021less, liu2019use} and evaluate Clover on the paragraph-to-video retrieval, where text sentences for each video are concatenated together as one text query. We do not use the ground-truth proposal for fair comparison with previous works;  (b) \textbf{Multiple-choice QA} on TGIF-Action \cite{jang2017tgif},  TGIF-Transition \cite{jang2017tgif}, MSRVTT-MC \cite{yu2018joint} and LSMDC-MC \cite{torabi2016learning}; (c) \textbf{Open-Ended QA} on TGIF-Frame~\cite{jang2017tgif}, MSRVTT-QA \cite{xu2017video}, MSVD-QA \cite{xu2017video} and LSMDC-FiB \cite{maharaj2017dataset}. For retrieval tasks, we only use the two uni-modal encoders of Clover for fine-tuning and inference. We adopt the two encoders to get the video and text embeddings, and calculate their cosine similarity for retrieval. For QA tasks, we use all the modules for fine-tuning and inference. More details on these datasets and their evaluation usage are provided in the 
supplementary material.

\begin{table*}[t]
\centering
\resizebox{0.80\textwidth}{!}{
\begin{tabular}{l|c|ccc|cc|cc|c}
\toprule
\multicolumn{1}{l|}{\multirow{2}{*}{Method}} & \multirow{2}{*}{Pre-training dataset} & \multicolumn{3}{c|}{TGIF} & \multicolumn{2}{c|}{MSRVTT} & \multicolumn{2}{c|}{LSMDC} & MSVD \\ \cline{3-10} 
\multicolumn{1}{c|}{} &  & Action & Transition & Frame & MC & QA & MC & FiB & QA \\ \midrule
clipBert \cite{lei2021less} & COCO, VG & 82.8 & 87.8 & 60.3 & 88.2 & 37.4 & - & - & - \\
JuskAsk \cite{yang2021just} & HTVQA69M\cite{yang2021just} & - & - & - & - & 41.5 & - & - & 46.3 \\
ALPRO \cite{li2022alignandprompt} & W2M+C3M & - & - & - & - & 42.1 & - & - & 45.9 \\
All-in-1 \cite{wang2022all} & W2M+H100M & 92.7 & 94.3 & 64.2 & 92.0 & 42.9 & 83.1 & - & 47.9 \\
VIOLET \cite{fu2021violet} & W2M+C3M+Y180M & 92.5 & 95.7 & 68.9 & 91.9 & 43.9 & 82.8 & 53.7 & 47.9 \\
MERLOT \cite{zellers2021merlot} & Y180M & 94.0 & 96.2 & 69.5 & 90.9 & 43.1 & 81.7 & 52.9 & - \\ \midrule
Clover (ours) & W2M+C3M & \textbf{95.0} & \textbf{98.2} & \textbf{71.6} & \textbf{95.2} & \textbf{44.1} & \textbf{83.7} & \textbf{54.1} & \textbf{52.4} \\ \bottomrule
\end{tabular}}
\caption{Performance comparison on transferring to downstream video question answering tasks.}
\vspace{-0.4cm}
\label{tab:vqa_sota}
\end{table*}

\begin{table*}[t]
\centering
\resizebox{0.90\textwidth}{!}{
\begin{tabular}{l|cccc|cccc|c|c}
\toprule
\multicolumn{1}{l|}{\multirow{2}{*}{Method}} & \multicolumn{4}{c|}{DiDeMo} & \multicolumn{4}{c|}{MSRVTT} & LSMDC-MC & TGIF-Frame \\ \cmidrule{2-11} 
\multicolumn{1}{c|}{} & R@1 & R@5 & R@10 & \multicolumn{1}{l|}{MedR} & R@1 & R@5 & \multicolumn{1}{l}{R@10} & \multicolumn{1}{l|}{MedR} & Acc & Acc \\ \midrule
Baseline  & 22.6 & 47.9 & 58.2 & 7 & 19.8 & 41.7 & 51.1 & 10 & 78.8 & 68.9 \\
+ TMA  & 24.7 & 49.7 & 60.0 & 6 & 22.7 & 42.2 & 52.2 & 9 & 80.0 & 69.3 \\
+ TMA+SM  & 25.3 & 49.5 & 60.5 & 6 & 22.5 & 42.7 & 52.0 & 9 & 80.5 & 69.4 \\
+ TMA+SM+RankL  & \textbf{26.4} & \textbf{51.1} & \textbf{61.3} & \textbf{5} & \textbf{23.4} & \textbf{43.3} & \textbf{52.4} & \textbf{9} & \textbf{80.7} & \textbf{69.7} \\ \bottomrule
\end{tabular}}
\caption{Effects of our pre-training tasks for Clover. We report \emph{zero-shot} text-video retrieval performance on DiDeMo and MSRVTT, and \emph{fine-tune} video QA performance on LSMDC-MC and TGIF-Frame. TMA, SM and RankL:  tri-modal alignment, semantic masking strategy and pair-wise ranking loss. }
\vspace{-0.2cm}
\label{tab:ablation}
\end{table*}

\begin{table}[]
\centering
\resizebox{0.98\linewidth}{!}{
\begin{tabular}{l|c|c}
\toprule
Method & \begin{tabular}[c]{@{}c@{}}Averaged \textbf{similarity scores} \\ of positive pairs\end{tabular} & \begin{tabular}[c]{@{}c@{}}Averaged \textbf{similarity margin} between \\ positive and negative samples\end{tabular} \\ \midrule
w/o TMA & 0.56 & 0.30 \\ 
w TMA & 0.65 & 0.35 \\ \bottomrule
\end{tabular}}
\caption{Affect of TMA on the video and text representations in learned embedding space. The experiment is conducted on MSRVTT-1kA test set under zero-shot setting.}
\vspace{-0.5cm}
\label{tab:avg cos_sim}
\end{table}

\noindent  \textbf{Implementation details}.
\label{sec:implementation details}
Following \cite{fu2021violet}, we initialize the video encoder with a VideoSwin-Base \cite{liu2022video} model pre-trained on Kinetics-400 \cite{kay2017kinetics}. The text encoder is initialized from pre-trained Bert-Base~\cite{devlin2019bert}. The multi-modal encoder is initialized from the first three layers of the pre-trained Bert-Base model. We train Clover end-to-end   during both pre-training and fine-tuning. We pre-train Clover for 40 epochs, using a batch size of 1024 on 64 NVIDIA A100 GPUs. We use AdamW \cite{loshchilov2018decoupled} optimizer with a weight decay 0.005 and betas (0.9, 0.98). The learning rate is first warmed-up by 4 epochs to 5e-5 and then decays following a cosine annealing decay schedule. We resize all the video frames to 224 $\times$ 224 and split each frame into patches with a size of 32 $\times$ 32. We choose hyper-parameter $\tau=0.05$, $\lambda=5$ and $\gamma=2$. We set dimension $D=768$ in our model. For each video, we randomly sample 8 frames while preserving their order in-between. For fine-tuning on retrieval tasks, we only fine-tune the uni-modal encoders of Clover with InfoNCE \cite{van2018representation} loss. For fine-tuning on video QA task, we add a simple MLPs that takes the multi-modal [CLS] embedding as input for classification, and optimize the whole Clover model with cross-entropy loss. During fine-tuning, we follow the conventional set-up in \cite{fu2021violet,luo2021clip4clip}, and all the fine-tuning experiments are performed on 8 NVIDIA A100 GPUs.\footnote{We provide our codes in the supplementary material and will release the codes to facilitate reproduction of our work.}

\begin{table*}[h]
\centering
\resizebox{0.90\linewidth}{!}{
\begin{tabular}{l|cc|ccc|c|c|c|c}
\toprule
\multirow{2}{*}{Method} & \multicolumn{2}{c|}{Model Architecture} & \multicolumn{3}{c|}{Training Objectives} & MSRVTT & DiDeMo & TGIF-Frame & LSMDC-MC \\ \cmidrule{2-10} 
 & \multicolumn{1}{c|}{\begin{tabular}[c]{@{}c@{}}Two uni-modal \\ encoder\end{tabular}} & \begin{tabular}[c]{@{}c@{}}Cross-modal \\ encoder\end{tabular} & \multicolumn{1}{c|}{InfoNCE} & \multicolumn{1}{c|}{MLM} & \begin{tabular}[c]{@{}c@{}}TMA +\\  Rankloss\end{tabular} & R@1 & R@1 & Acc & Acc \\ \midrule
IND-A & \multicolumn{1}{c|}{\Checkmark} &  & \multicolumn{1}{c|}{\Checkmark} & \multicolumn{1}{c|}{} &  & 20.7 & 22.9 & - & - \\
IND-F & \multicolumn{1}{c|}{\Checkmark} & \Checkmark & \multicolumn{1}{c|}{} & \multicolumn{1}{c|}{\Checkmark} &  & - & - & 69.0 & 79.0 \\
COMB & \multicolumn{1}{c|}{\Checkmark} & \Checkmark & \multicolumn{1}{c|}{\Checkmark} & \multicolumn{1}{c|}{\Checkmark} &  & 19.8 & 22.6 & 68.9 & 78.8 \\
Clover & \multicolumn{1}{c|}{\Checkmark} & \Checkmark & \multicolumn{1}{c|}{} & \multicolumn{1}{c|}{\Checkmark} & \Checkmark & \textbf{23.4} & \textbf{26.4} & \textbf{69.7} & \textbf{80.7} \\ \bottomrule
\end{tabular}}
\caption{Comparisons with task-independently trained models (\emph{IND}) and naive combination of the uni- and cross- modal encoder (\emph{COMB})}
\vspace{-0.3cm}
\label{tab:ablation_comb_ind}
\end{table*}

\subsection{Comparing to State-of-the-art}
\noindent  \textbf{Text-to-video retrieval}. 
Tab.~\ref{tab:retrieval_SOTA} shows the retrieval results on the DiDeMo \cite{anne2017localizing}, MSRVTT \cite{xu2016msr} and LSMDC \cite{maharaj2017dataset} datasets with both zero-shot and fine-tuning settings. Our Clover surpasses previous works on all the datasets by a large margin. The significant performance gain under zero-shot evaluation demonstrates its stronger generalization ability. Specifically, Clover brings $4.9\%$ gain in R@10 over MCQ \cite{ge2022bridging},  the current SOTA method on retrieval tasks with the  same pre-train data, averaged on the three datasets. Moreover, Clover outperforms VIOLET~\cite{fu2021violet}, though VIOLET adopts a considerably larger pre-training  dataset (YTT180M \cite{zellers2021merlot} of 180M visual-text pairs vs.\ WebVid2M+CC3M of only 5.5M samples).  Clover also presents superior  performance when fine-tuned on the downstream datasets, compared with the SOTAs. On all three datasets, Clover surpasses previous works by a large margin across all the metrics and obtains an average improvement of $8.7\%$ on R@10. Moreover, different from some  previous work~\cite{lei2021less, fu2021violet, li2022alignandprompt} that adopts a joint multi-modal encoder and  requires exhaustively pairing every video with every text during retrieval, Clover directly deploys the video encoder and text encoder for retrieval tasks,  making it  much more efficient.

\noindent  \textbf{Video question answering}.
Tab.~\ref{tab:vqa_sota} reports the results of Clover and current   SOTAs on four video QA datasets. Though using much less pre-training data, Clover outperforms another unified VidL model \emph{i.e.,} VIOLET \cite{fu2021violet}  significantly. Moreover, compared to the current SOTA method MERLOT\cite{zellers2021merlot}, Clover brings  improvements of 1.0\% on TGIF-Action, 2.0\% on TGIF-Transition, 2.1\% on TGIF-Frame, 4.3\% on MSRVTT-MC, 1.0\% on MSRVTT-QA, 2.0\% on LSMDC-MC and 1.2\% on LSMDC-FiB. Note that JustAsk \cite{yang2021just} and MERLOT~\cite{zellers2021merlot} are specifically designed for video QA and trained on orders of magnitude larger datasets (\emph{e.g.,} HTVQA69M \cite{yang2021just}, YTT180M \cite{zellers2021merlot}). The experiment results again confirm the superiority of Clover.

\subsection{Analysis}
We conduct ablation experiments on two retrieval datasets (DiDeMo and MSRVTT) and two Video QA datasets (TGIF-Frame and LSMDC-MC). Due to the computational resource limit, we randomly sample 1 million video-text pairs  from WebVid2M \cite{bain2021frozen}    to build the WebVid-1M subset  for model pre-training under   all the ablation studies. We keep the model architecture unchanged but pre-train the model with only MLM and InfoNCE losses as a baseline, which  is similar to the baseline adopted in VIOLET \cite{fu2021violet}. The only difference is that we replace the ITM applied to the cross-modal encoder outputs in VIOLET \cite{fu2021violet} with InfoNCE applied to the outputs of uni-modal encoders and separate the text encoder from the cross-modal encoder. All the ablation experiments are conducted with batch size 1024 on 32 NVIDIA A100 GPUs.


\noindent  \textbf{Effect of tri-modal alignment}. Tri-modal alignment (TMA) aims to better correlate cross-modal alignment and   fusion. To evaluate the contribution of TMA, we remove the pair-wise ranking objective and semantic masking strategy, and employ the classical MLM task as in the baseline method. Compared with the baseline, TMA explicitly associates the outputs of uni-modal encoders with the outputs of the multi-modal encoder. As shown in Tab.~\ref{tab:ablation}, the model trained with TMA outperforms the baseline on both retrieval and video QA, demonstrating its effectiveness. For a better understanding of the effect of TMA, we further report averaged cosine similarity of videos and texts in MSRVTT dataset (in zero-shot setting) calculated by models trained with/without TMA, respectively.  For a fair comparison, we select all video queries (197 in total) that were successfully responded to by both models. We recall 100 results for each query, and take the ground-truth pairs as positive and others as negative. 
As shown in  Tab.~\ref{tab:avg cos_sim}, the model armed with tri-modal alignment assigns higher similarity scores to positive pairs compared to the model without it. Meanwhile, with  tri-modal alignment, the margin between the positive and negative samples is also increased. It reveals that with the help of tri-modal alignment, the distance between video and text that contain consistent semantics is  reduced, while the margin between the misaligned samples is increased. The results demonstrate that with fused multi-modal representation as anchors, tri-modal alignment helps the video and language modalities to be better aligned.

\begin{figure}[tb]
  \centering
  \includegraphics[width=1\columnwidth]{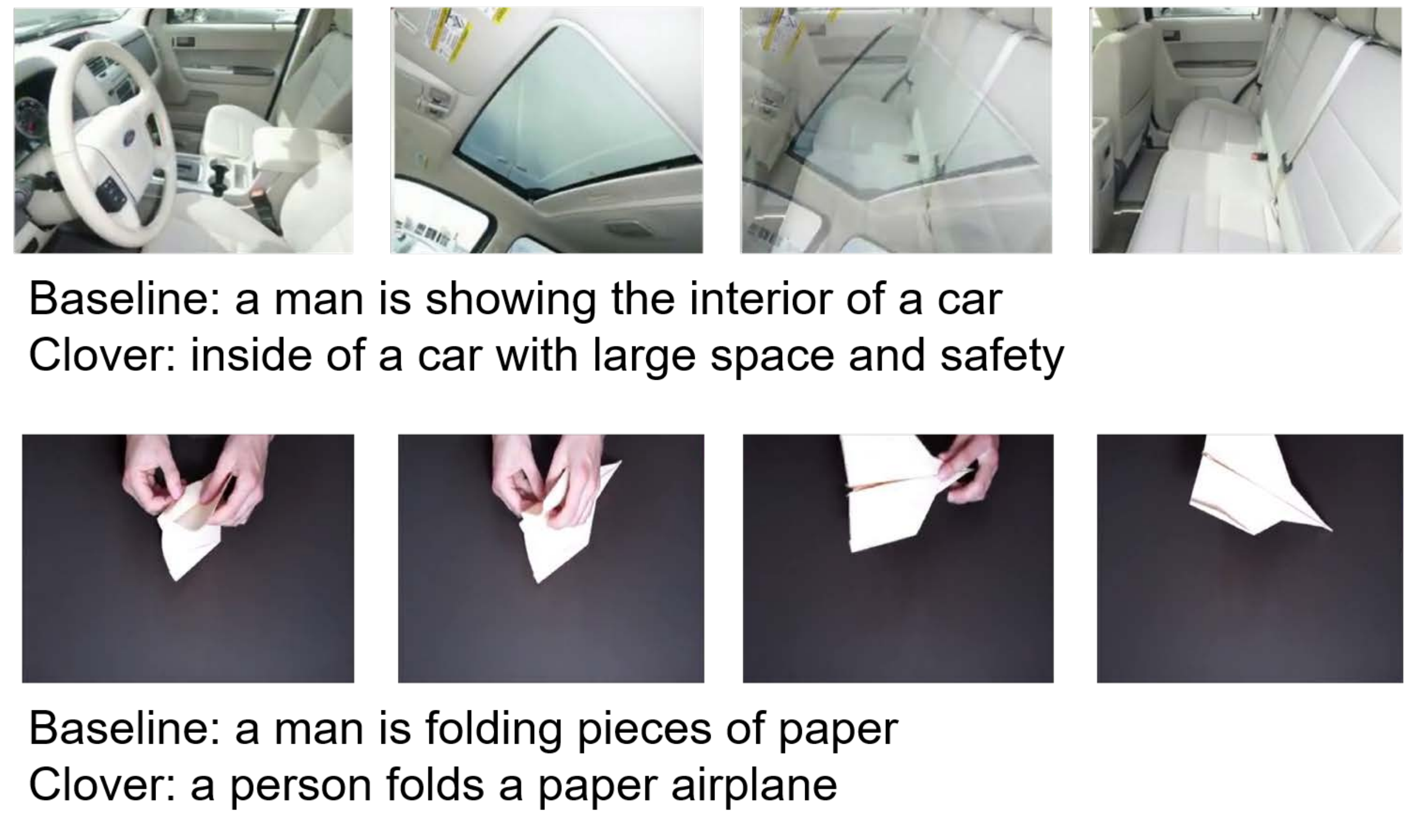}
 \vspace{-0.3cm}
  \caption{Qualitative results of zero-shot video to text retrieval results on MSRVTT \cite{xu2016msr}.}
   \label{fig:vis_ret}
   \vspace{-1\baselineskip}
\end{figure}

\noindent \textbf{Effect of semantic masking strategy}. We incorporate a semantic masking strategy to form the masked pairs in tri-modal alignment task.  As shown in Tab.~\ref{tab:ablation}, on more diverse evaluation benchmarks, e.g. DiDemo,  where each video is paired with 5 different ground-truth texts, using masked samples brings more gains. The result reveals that the proposed semantic masking strategy facilitates the model to capture key semantic information in more complex scenes and makes it achieve better results.

\noindent \textbf{Effect of pair-wise ranking}. To make the model capture the semantic information difference between the masked pairs and complete pairs, we adopt the pair-wise rank loss. As shown in Tab.~\ref{tab:ablation}, with pair-wise ranking loss, the performance of the model is further improved. We also show the visualization results below to further illustrate the effect of pair-wise ranking.

\noindent \textbf{Clover makes cross-modal alignment and fusion mutually improving}. We compare Clover with the ones that use the same architecture but employ different pre-training tasks in Tab.~\ref{tab:ablation_comb_ind}. \emph{IND-A} represents the model only trained with InfoNCE loss for cross-modal alignment. \emph{IND-F} indicates the model trained with MLM loss for cross-modal fusion. We also report results achieved by the model (\emph{i.e., COMB}) that simply combines the uni-modal encoders and multi-modal encoder as well as the training objectives. We can see that the simple combination   hurts the performance, while our Clover achieves superior performance than the task-independently trained \emph{IND-A} and \emph{IND-F}. It reveals that Clover is able to get the model's cross-modal alignment and fusion capability mutually enhanced. 

\begin{figure}[tb]
  \centering
  \includegraphics[width=1\columnwidth]{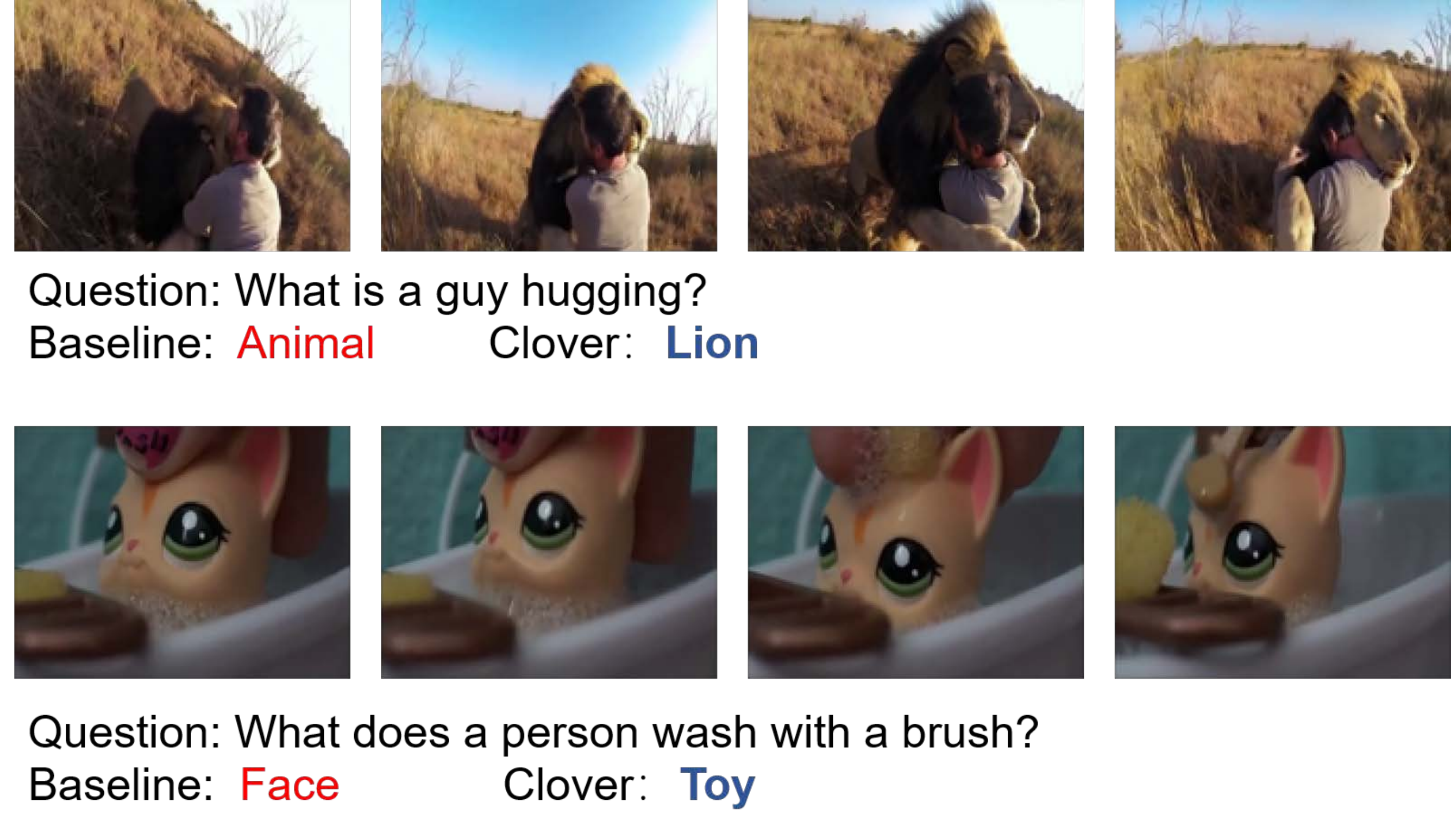}
  \caption{Qualitative results of video question answering results on MSRVTT-QA \cite{xu2017video}.}
   \label{fig:vis_vqa}
   \vspace{-1\baselineskip}
\end{figure}

\noindent \textbf{Qualitative analysis}. 
In Fig.~\ref{fig:vis_ret} and Fig.~\ref{fig:vis_vqa}, we show the qualitative results of Clover and the baseline method on zero-shot video-text retrieval and VQA tasks. Specifically, we present the query videos with the matched texts, and the video-question pair with the answer. Clover empowers the model with stronger video-text understanding capability. 
For example, for the second video in Fig.~\ref{fig:vis_ret}, Clover returns the text with ``paper airplane", which is a more accurate description of the video content; for the first video in Fig.~\ref{fig:vis_vqa}, Clover generates a more accurate answer ``Lion" than the baseline answer ``Animal".
The results demonstrate the superiority of Clover.

\vspace{-0.2cm}
\section{Conclusion} \label{sec:conclusion}

In this paper, we present Clover, a new end-to-end Video-Language pre-training method for both high-efficiency video-text retrieval and video question answering. Clover introduces a novel Tri-modal Alignment task to better  align the representations from visual, text and fused modalities, which explicitly correlate the uni-modal encoder and multi-modal encoder.  It also introduces semantic masking strategy and pair-wise ranking loss to further improve the cross-modality modal training.  Extensive experiments conducted on the three retrieval datasets and eight video QA datasets clearly demonstrated, as a general video-text model, its   consistent superiority  for video-text understanding.

{\small
\bibliographystyle{ieee_fullname}
\bibliography{egbib}
}

\end{document}